\def\year{2020}\relax
\documentclass[letterpaper]{article} 
\usepackage{aaai20}  
\usepackage{times}  
\usepackage{helvet} 
\usepackage{courier}  
\usepackage[hyphens]{url}  
\usepackage{graphicx} 
\usepackage{graphicx}  
\usepackage{mathrsfs,amssymb,mathtools}
\usepackage{algorithm}
\usepackage{subfigure}
\usepackage{verbatim}
\usepackage{enumerate}
\usepackage{bm}
\usepackage{soul}
\usepackage[noend]{algpseudocode}
\usepackage[figuresright]{rotating}
\usepackage{multirow}
\usepackage{booktabs}

\title{Hierarchical Graph Pooling with Structure Learning}

\author{Zhen Zhang\textsuperscript{\rm 1}, Jiajun Bu\textsuperscript{\rm 1}, Martin Ester\textsuperscript{\rm 2}, Jianfeng Zhang\textsuperscript{\rm 3},\\ \Large \textbf{Chengwei Yao\textsuperscript{\rm 1}, Zhi Yu\textsuperscript{\rm 1}, Can Wang\textsuperscript{\rm 1}}\\
\textsuperscript{\rm 1}Zhejiang University, \textsuperscript{\rm 2}Simon Fraser University, \textsuperscript{\rm 3}Alibaba Group\\
zhen\_zhang@zju.edu.cn
}
\begin{document}

\maketitle

\begin{abstract}
Graph Neural Networks (GNNs), which generalize deep neural networks to graph-structured data, have drawn considerable attention and achieved state-of-the-art performance in numerous graph related tasks. However, existing GNN models mainly focus on designing graph convolution operations. The graph pooling (or downsampling) operations, that play an important role in learning hierarchical representations, are usually overlooked. In this paper, we propose a novel graph pooling operator, called Hierarchical Graph Pooling with Structure Learning (HGP-SL), which can be integrated into various graph neural network architectures. HGP-SL incorporates graph pooling and structure learning into a unified module to generate hierarchical representations of graphs. More specifically, the graph pooling operation adaptively selects a subset of nodes to form an induced subgraph for the subsequent layers. To preserve the integrity of graph's topological information, we further introduce a structure learning mechanism to learn a refined graph structure for the pooled graph at each layer. By combining HGP-SL operator with graph neural networks, we perform graph level representation learning with focus on graph classification task. Experimental results on six widely used benchmarks demonstrate the effectiveness of our proposed model.
\end{abstract}

\section{Introduction}
Deep neural networks with convolution and pooling layers have achieved great success in various challenging tasks, ranging from computer vision \cite{he2016deep}, natural language understanding \cite{bahdanau2014neural} to video processing \cite{karpathy2014large}. The data in these tasks are typically represented in the Euclidean space (i.e., modeled as 2-D or 3-D tensors), thus usually containing locality and order information for the convolution operations \cite{defferrard2016convolutional}. However, in many real-world problems, a large amount of data, such as social networks, chemical molecules and biological networks, are lying on non-Euclidean domains that can be naturally represented as graphs. Due to the neural network's powerful capabilities, it's quite appealing to generalize the convolution and pooling operations to graph-structured data.

Recently, there have been a myriad of attempts to generalize the convolution operations to arbitrary graphs, referred to as graph neural networks (GNNs for short). In general, these algorithms can be classified into two big categories: spectral and spatial approaches. For the spectral methods, they typically define the graph convolution operations based on graph Fourier transform \cite{bruna2013spectral,defferrard2016convolutional,kipf2016semi}. For the spatial methods, the graph convolution operations are devised by aggregating the node representations directly from its neighborhood \cite{hamilton2017inductive,monti2017geometric,velivckovic2017graph,morris2019weisfeiler}. Majority of the aforementioned methods mainly involve transforming, propagating and aggregating node features across the graph, which can fit in the message passing scheme \cite{gilmer2017neural}. GNNs have been applied to different types of graphs \cite{velivckovic2017graph,derr2018signed}, and obtained outstanding performance in numerous graph related tasks, including node classification \cite{kipf2016semi}, link prediction \cite{schlichtkrull2018modeling,zhang2018anrl} and recommendation \cite{ying2018graph}, {\it etc}.

Nevertheless, the pooling operations in graphs have not been extensively studied yet, though they act a pivotal part in learning hierarchical representations for the task of graph classification \cite{ying2018hierarchical}. The goal of graph classification is to predict the label associated with the entire graph by utilizing its node features and graph structure information, i.e., a graph level representation is needed. GNNs are originally designed to learn meaningful node level representations, thus a commonly adopted approach to generate graph level representation is to globally summarize all the node representations in the graph. Although workable, the graph level representation generated via this way is inherently ``flat", since the entire graph structure information is neglected during this process. Furthermore, GNNs can only pass messages between nodes through edges, but cannot aggregate node information in a hierarchical way. Meanwhile, graphs often have different substructures and nodes are of different roles, therefore they should contribute differently to the graph level representation. For example, in the protein-protein interaction graphs, the certain substructures may represent some specific functionalities, which are of great significance to predict the whole graph characteristics. To capture both the graph's local and global structure information, a hierarchical pooling process is demanded.

There exists some very recent work that focuses on the hierarchical pooling procedure in GNNs \cite{ying2018hierarchical,gao2019graph,diehl2019edge,gao2019learning}. These models usually coarsen the graphs through grouping or sampling nodes into subgraphs level by level, thus the entire graph information is gradually reduced to the hierarchical induced subgraphs. However, the graph pooling operations still have room for improvement. In node grouping approaches, the hierarchical pooling methods \cite{ying2018hierarchical,diehl2019edge} suffer from high computational complexity, which require additional neural networks to downsize the nodes. In node sampling approaches, the generated induced subgraph \cite{gao2019graph,lee2019self} might fail to preserve the key substructures and eventually lose the completeness of graph topological information. For instance, two nodes that are not directly connected but sharing many common neighbors in the original graph might become unreachable from each other in the induced subgraph, even if intuitively they ought to be ``close" in the subgraph. Therefore, the distorted graph structure will hinder the message passing in subsequent layers.

To address the aforementioned limitations, we propose a novel graph pooling operator HGP-SL to learn hierarchical graph level representations. Specifically, HGP-SL first adaptively selects a subset of nodes according to our defined node information score, which fully utilizes both the node features and graph topological information. In addition, the proposed graph pooling operation is a non-parametric step, therefore no additional parameters need to be optimized during this procedure. Then, we apply a structure learning mechanism with sparse attention \cite{martins2016softmax} to the pooled graph, aiming to learn a refined graph structure that preserves the key substructures in the original graph. We integrate the pooling operator into graph convolutional neural network to perform graph classification and the whole procedure can be optimized in an end-to-end manner. To summarize, the main contributions of this paper are as follows:
\begin{itemize}
	\item We introduce a novel graph pooling operator HGP-SL that can be integrated into various graph neural network architectures. Similarly to the pooling operations in convolutional neural networks, our proposed graph pooling operation is non-parametric\footnote{Note that the pooling process itself is non-parametric, however the structure learning mechanism indeed has an attention parameter. Thus, the overall HGP-SL operator is not non-parametric.} and very easy to implement. 
	\item To the best of our knowledge, we are the first to design a structure learning mechanism for the pooled graph, which has the advantage of learning a refined graph structure to preserve the graph's key substructures.
	\item We conduct extensive experiments on six public datasets to demonstrate HGP-SL's effectiveness as well as superiority compared to a range of state-of-the-art methods.
\end{itemize}

\section{Related Work}
\subsection{Graph Neural Networks}
GNNs can be generally categorized into two branches: spectral and spatial approaches. The spectral methods typically define the parameterized filters according to graph spectral theory. \cite{bruna2013spectral} first proposed to define convolution operations for graph in the Fourier transform domain. Due to its heavy computation cost, it has difficulty in scaling to large graphs. Later on, \cite{defferrard2016convolutional} improved its efficiency by approximating the K-polynomial filters through Chebyshev expansion. GCN \cite{kipf2016semi} further simplified the ChebNet by truncating the Chebyshev polynomial to the first-order approximation of the localized spectral filters.

The spatial approaches design convolution operations by directly aggregating the node's neighborhood information. Among them, GraphSAGE \cite{hamilton2017inductive} proposed an inductive algorithm that can generalize to unseen nodes by aggregating its neighborhood content information. GAT \cite{velivckovic2017graph} utilized attention mechanism to aggregate nodes' neighborhood representations with different weights. JK-Net \cite{xu2018representation} leveraged flexible neighborhood ranges to enable better node representations. More details can be found in several comprehensive surveys on graph neural networks \cite{zhou2018graph,zhang2018deep,wu2019comprehensive}. Nevertheless, the above mentioned two branches of GNNs are mainly designed for learning meaningful node representations, and unable to generate hierarchical graph representations due to the lack of pooling operations.

\subsection{Graph Pooling}
Pooling operations in GNNs can scale down the size of inputs and enlarge the receptive fields, thus giving rise to better generalization and performance. DiffPool \cite{ying2018hierarchical} proposed to softly assign nodes to a set of clusters using neural networks, which forms a dense cluster assignment matrix and is computation expensive. gPool \cite{gao2019graph} and SAGPool \cite{lee2019self} devised a top-K node selection procedure to form an induced subgraph for the next input layer. Though efficient, it might lose the completeness of the graph structure information and result in isolated subgraphs, which will hamper the message passing process in subsequent layers. EdgePool \cite{diehl2019edge} designed pooling operation by contracting the edges in the graph, but its flexibility is poor because it will always pool roughly half of the total nodes. iPool \cite{gao2019ipool} presented a parameter-free pooling scheme which is invariant to graph isomorphism. EigenPool \cite{ma2019graph} introduced a pooling operator based on the graph Fourier transform, which controls the pooling ratio through spectral clustering and it's also very time consuming.

In addition, there are also some approaches that perform global pooling. For instance, Set2Set \cite{vinyals2015order} implemented the global pooling operation by aggregating information through LSTMs \cite{hochreiter1997long}. DGCNN \cite{zhang2018end} pooled the graph according to the last channel of the feature map values which are sorted in the descending order. Graph topological based pooling operations are proposed in \cite{defferrard2016convolutional} and \cite{rhee2017hybrid} as well, where Graclus method \cite{dhillon2007weighted} is employed as a pooling module. 

\section{The Proposed Model}
\subsection{Notations and Problem Formulation}
Given a set of graph data $G = \{\mathcal{G}_1, \mathcal{G}_2, \cdots, \mathcal{G}_n\}$, where the number of nodes and edges in each graph might be quite different. For an arbitrary graph $\mathcal{G}_i = (\mathcal{V}_i, \mathcal{E}_i, \mathbf{X}_i)$, we have $n_i$ and $e_i$ denote the number of nodes and edges, respectively. Let $\mathbf{A}_i \in \mathbb{R}^{n_i \times n_i}$ be the adjacent matrix describing its edge connection information and $\mathbf{X}_i \in \mathbb{R}^{n_i \times f}$ represents the node feature matrix, where $f$ is the dimension of node attributes. Label matrix $\mathbf{Y} \in \mathbb{R}^{n \times c}$ indicates the associated labels for each graph, i.e., if $\mathcal{G}_i$ belongs to class $j$, then $\mathbf{Y}_{ij} = 1$, otherwise $\mathbf{Y}_{ij} = 0$. Since the graph structure and node numbers change between layers due to the graph pooling operation, we further represent the $i$-th graph fed into the $k$-th layer as $\mathcal{G}_i^{k}$ with $n_i^k$ nodes. The adjacent matrix and hidden representation matrix are then denoted as $\mathbf{A}_i^k \in \mathbb{R}^{n_i^k \times n_i^k}$ and $\mathbf{H}_i^k \in \mathbb{R}^{n_i^k \times d}$. With the above notations, we formally define our problem as follows:

\textbf{Input:}
Given a set of graphs $G_L$ with its label information $\mathbf{Y}_L$, the number of graph neural network layers $K$, pooling ratio $r$, and representation dimension $d$ in each layer.

\textbf{Output:} Our goal is to predict the unknown graph labels of $G/G_L$ with graph neural network in an end-to-end way.

\subsection{Graph Convolutional Neural Network}
Graph convolutional neural network (or GCN) \cite{kipf2016semi} has shown to be very efficient and achieved promising performance in various challenging tasks. Thus, we choose GCN as our model's building block and briefly review its mechanism in this subsection. Please note that our proposed HGP-SL operator can also be integrated into other graph neural network architectures like GraphSAGE \cite{hamilton2017inductive} and GAT \cite{velivckovic2017graph}. We will discuss this in the experiment section. For the $k$-th layer in GCN, it takes graph $\mathcal{G}$'s adjacent matrix $\mathbf{A}$ and hidden representation matrix $\mathbf{H}_k$ as input, then the next layer's output will be generated as follows:
\begin{equation}
	\mathbf{H}_{k+1} = \sigma(\tilde{\mathbf{D}}^{-\frac{1}{2}}\tilde{\mathbf{A}}\tilde{\mathbf{D}}^{-\frac{{1}}{2}}\mathbf{H}_k\mathbf{W}^k),
	\label{eq:conv}
\end{equation}
where $\sigma(\cdot)$ is the non-linear activation function and $\mathbf{H}_0 = \mathbf{X}$, $\tilde{\mathbf{A}} = \mathbf{A} + \mathbf{I}$ is the adjacent matrix with self-connections. $\tilde{\mathbf{D}}$ is the diagonal degree matrix of $\tilde{\mathbf{A}}$, and $\mathbf{W}^k \in \mathbb{R}^{d_k \times d_{k+1}}$ is a trainable weight matrix. For the ease of parameter tuning, we set output dimension $d_{k+1} = d_k = d$ for all layers.

\subsection{The Overall Neural Network Architecture}
\begin{figure*}
  \centering
  \includegraphics[width=0.8\textwidth]{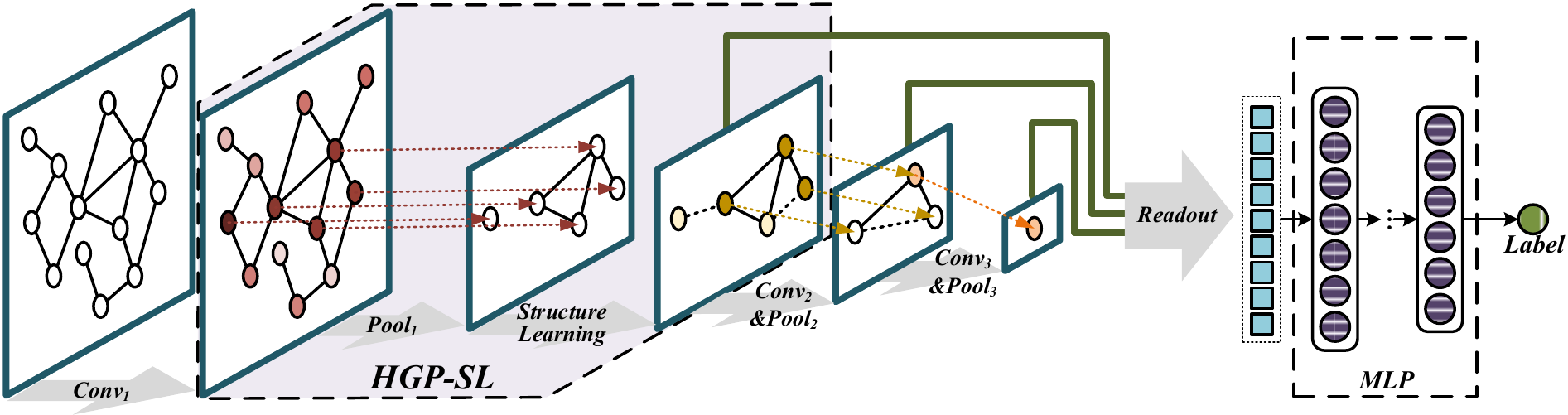}
  \caption{Architecture of proposed HGP-SL operator combined with graph neural network. The dashed box demonstrates the workflow of HGP-SL, which involves graph pooling and structure learning. The learned edges are represented as dashed lines in the graph. This procedure (convolution and pooling operations) is repeated several times. Then, a readout function is applied to aggregate node representations to make a fixed size representation, which goes through MLP layers for graph classification.}
  \label{fig:model}
\end{figure*}
Figure \ref{fig:model} provides an overview of our proposed $\underline{\textbf{H}}$ierarchical $\underline{\textbf{G}}$raph $\underline{\textbf{P}}$ooling with $\underline{\textbf{S}}$tructure $\underline{\textbf{L}}$earning (HGP-SL) that combines with graph neural network, where graph pooling operations are added between graph convolution operations. The proposed HGP-SL operator is composed of two major components: 1) graph pooling, which preserves a subset of informative nodes and forms a smaller induced subgraph; and 2) structure learning, which learns a refined graph structure for the pooled subgraph. The advantage of our proposed structure learning lies in its capability to preserve the essential graph structure information, which will facilitate the message passing procedure. As in this illustrative example, the pooled subgraph might exist isolated nodes but intuitively ought to be connected, thus it would hinder the information propagation in subsequent layers especially when aggregating information from its neighborhood nodes. The whole architecture is the stacking of convolution and pooling operations, thus making it possible to learn graph representations in a hierarchical way. Then, a readout function is utilized to summarize node representations in each level, and the final graph level representation is the addition of different levels' summarizations. At last, the graph level representation is fed into a Multi-Layer Perceptron (MLP) with softmax layer to perform graph classification task. In what follows, we give the details of graph pooling and structure learning layers.

\subsection{Graph Pooling Operation}
In this subsection, we introduce our proposed graph pooling operation to enable down-sampling on graph data. Inspired by \cite{gao2019graph,lee2019self,gao2019ipool}, the pooling operation identifies a subset of informative nodes to form a new but smaller graph. Here, we design a non-parametric pooling operation, which can fully utilize both the node features and graph structure information. 

The key of our proposed graph pooling operation is to define a criterion that guides the node selection procedure. To perform node sampling, we first introduce a criterion named node information score to evaluate the information that each node contains given its neighborhood. Generally, if a node's representation can be reconstructed by its neighborhood representations, it means this node can probably be deleted in the pooled graph with almost no information loss. Here, we formally define the node information score as the Manhattan distance between the node representation itself and the one constructed from its neighbors:
\begin{equation}
	\mathbf{p} = \gamma(\mathcal{G}_i) = \Vert(\mathbf{I}_i^k - (\mathbf{D}_i^k)^{-1}\mathbf{A}_i^k)\mathbf{H}_i^k\Vert_1,
	\label{eq:rank}
\end{equation}  
where $\mathbf{A}_i^k \in \mathbb{R}^{n_i^k \times n_i^k}$ and $\mathbf{H}_i^k \in \mathbb{R}^{n_i^k \times d}$ are the adjacent and node representations matrices. $\Vert\cdot\Vert_1$ performs $\ell_1$ norm row-wisely. $\mathbf{D}_i^k$ represents the diagonal degree matrix of $\mathbf{A}_i^k$, and $\mathbf{I}_i^k$ is the identity matrix. Therefore, we have $\mathbf{p} \in \mathbb{R}^{n_i}$ encode the information score of each node in the graph.

After having obtained the node information score, we can now select nodes that should be preserved by the pooling operator. To approximate the graph information, we choose to preserve the nodes that can not be well represented by their neighbors, i.e., the nodes with relative larger node information score will be preserved in the construction of the pooled graph, because they can provide more information. In details, we first re-order the nodes in graph according to their node information scores, then a subset of top-ranked nodes are selected as follows:
\begin{eqnarray}
	&{\rm idx} = \text{top-rank}(\mathbf{p}, \lceil{r*n_i^k}\rceil) \nonumber \\
	&\tilde{\mathbf{H}}_i^{k+1} = \mathbf{H}_i^k({\rm idx},:) \\
	&\mathbf{A}_i^{k+1} = \mathbf{A}_{i}^k({\rm idx}, {\rm idx}), \nonumber
\end{eqnarray}
where $r$ is the pooling ratio and top-rank$(\cdot)$ denotes the function that returns the indices of the top $n_i^{k+1}=\lceil r*n_i^k \rceil$ values. $\mathbf{H}_i^k({\rm idx},:)$ and $\mathbf{A}_i^k({\rm idx}, {\rm idx})$ perform the row or (and) column extraction to form the node representation matrix and adjacent matrix for the induced subgraph. Thus, we have $\tilde{\mathbf{H}}_i^{k+1} \in \mathbb{R}^{n_i^{k+1} \times d}$ and $\mathbf{A}_i^{k+1} \in \mathbb{R}^{n_i^{k+1} \times n_i^{k+1}}$ represent the node feature and graph structure information of next layer .

\subsection{Structure Learning Mechanism}
In this subsection, we present how our proposed structure learning mechanism learns a refined graph structure in the pooled graph. As we have illustrated in Figure \ref{fig:model}, the pooling operation might result in highly related nodes being disconnected in the induced subgraph, which loses the completeness of the graph structure information and further hinders the message passing procedure. Meanwhile, the graph structure obtained from domain knowledge (e.g., social network) or established by human (e.g., KNN graph) are usually non-optimal for the learning task in graph neural networks, due to the lost or noisy information. To overcome this problem, \cite{li2018adaptive} proposed to adaptively estimate graph Laplacian using an approximate distance metric learning algorithm, which might lead to local optimal solution. \cite{jiang2019semi} introduced to learn the constructed graph structure for node label estimation, however it generates dense connected graph and is not applicable in our hierarchical graph level representation learning scenario. 

Here, we develop a novel structure learning layer, which learns sparse graph structure through sparse attention mechanism \cite{martins2016softmax}. For graph $\mathcal{G}_i$'s pooled subgraph $\mathcal{G}_i^k$ at its $k$-th layer, we take its structure information $\mathbf{A}_i^k \in \mathbb{R}^{n_i^k \times n_i^k}$ and hidden representations $\mathbf{H}_i^k \in \mathbb{R}^{n_i^k \times d}$ as input. Our target is to learn a refined graph structure that encodes the underlying pairwise relationship between each pair of nodes. Formally, we utilize a single layer neural network parameterized by a weight vector $\stackrel{\rightarrow}{\mathbf{a}} \in \mathbb{R}^{1 \times 2d}$. Then, the similarity score between node $v_p$ and $v_q$ calculated by the attention mechanism can be expressed as:
\begin{equation}
	\mathbf{E}_i^k(p,q) = {\sigma}(\stackrel{\rightarrow}{\mathbf{a}}[\mathbf{H}_i^k(p,:)||\mathbf{H}_i^k(q,:)]^{\top}) + \lambda\cdot\mathbf{A}_i^k(p,q),
\end{equation}
where $\sigma(\cdot)$ is the activation function like ${\rm ReLU}(\cdot)$ and $||$ represents the concatenation operation. $\mathbf{H}_i^k(p,:) \in \mathbb{R}^{1 \times d}$ and $\mathbf{H}_i^k(q,:) \in \mathbb{R}^{1 \times d}$ indicate the $p$-th and $q$-th row of matrix $\mathbf{H}_i^k$, which denote the representations of node $v_p$ and $v_q$, respectively. Specifically, $\mathbf{A}_i^k$ encodes the induced subgraph structure information, where $\mathbf{A}_i^k(p,q) = 0$ if node $v_p$ and $v_q$ are not directly connected. We incorporate $\mathbf{A}_i^k$ into our structure learning layer to bias the attention mechanism to give a relatively larger similarity score between directly connected nodes, and at the same time try to learn the underlying pairwise relationships between disconnected nodes. $\lambda$ is a trade-off parameter between them.

To make the similarity score easily comparable across different nodes, we could normalize them across nodes using the softmax function:
\begin{equation}
\mathbf{S}_i^k(p,q) = \frac{{\rm exp}(\mathbf{E}_i^k(p,q))}{\sum_{m=1}^{n_i^k}{\rm exp}(\mathbf{E}_i^k(p,m))}.	
\label{eq:softmax}
\end{equation}
However, the softmax transformation always has non-zero values and thus results in dense fully connected graph, which may introduce lots of noise into the learned structure. Hence, we propose to utilize sparsemax function \cite{martins2016softmax}, which retains most the important properties of softmax function and has in addition the ability of producing sparse distributions. The ${\rm sparsemax}(\cdot)$ function aims to return the Euclidean projection of input onto the probability simplex and can be formulated as follows:
\begin{eqnarray}
	&\mathbf{S}_i^k(p,q) = {\rm sparsemax}(\mathbf{E}_i^k(p,q)) \nonumber \\
	&{\rm sparsemax}(\mathbf{E}_i^k(p,q)) = [\mathbf{E}_i^k(p,q) - \tau{(\mathbf{E}_i^k(p,:))}]_+,
	\label{eq:sparse_max}
\end{eqnarray}
where $[x]_+ = {\rm max}\{0, x\}$, and $\tau(\cdot)$ is the threshold function that returns a threshold according to the procedure shown in Algorithm \ref{algo:tau_calculation}. Thus, ${\rm sparsemax}(\cdot)$ preserves the values above the threshold and the other values will be truncated to zeros, which brings sparse graph structure. Similarly to softmax function, ${\rm sparsemax}(\cdot)$ also has the properties of non-negative and sum-to-one, that's to say, $\mathbf{S}_i^k(p,q) \geq 0$ and $\sum_{q=1}^{n_i^k}\mathbf{S}_i^k(p,q) = 1$. The proof procedure is available in the supplemental material.
\begin{algorithm}
  \caption{The calculation procedure of function $\tau(\cdot)$}
  \label{algo:tau_calculation}
  \begin{algorithmic}[1]
    \Require {input vector $\mathbf{z} \in \mathbb{R}^{n}$}.
    \State Sort $\mathbf{z}$ into $\mathbf{u}$: $u_1 \ge u_2 \ge \cdots \ge u_n$.
    \State Get $\rho = \max \{1 \le j \le n: u_j + \frac{1}{j}(1 - \sum_{i=1}^ju_i) > 0\}$.
    \State Define $\tau(\mathbf{z}) = \frac{1}{\rho}(\sum_{i=1}^{\rho}u_i - 1)$.
  \end{algorithmic}
\end{algorithm}

\subsection{Improving Structure Learning Efficiency}
For large scale graphs, it will be computation expensive to calculate the similarities between each pair of nodes during the learning of structure $\mathbf{S}_i^{k}$. If we further take graph's localization and smoothness properties into account, it is reasonable to constrain the calculation process within the node's $h$-hop neighborhood ($h=2$ or $3$). Therefore, the computation cost of $\mathbf{S}_i^{k}$ can be greatly reduced. 

\subsection{GCN and Graph Pooling Revisiting}
After having obtained the refined graph structure $\mathbf{S}_i^k$, we conduct graph convolution and pooling operations in the following layers based on $\tilde{\mathbf{H}}_i^k$ and $\mathbf{S}_i^k$ (instead of $\mathbf{A}_i^k$). Thus, Equation (\ref{eq:conv}) can be simplified as follows:
\begin{equation}
	\mathbf{H}_i^k=\sigma(\mathbf{S}_i^k\tilde{\mathbf{H}}_i^k\mathbf{W}^k).
\end{equation}
Since the learned $\mathbf{S}_i^k$ satisfies $\sum_{q=1}^{n_i^k}\mathbf{S}_i^k(p,q) = 1$, therefore we have the diagonal matrix $\mathbf{D}_i^k = {\rm Diag}(d_1,d_2,\cdots,d_{n_i^k})$ with $d_p = \sum_q^{n_i^k}{\mathbf{S}_i^k(p,q)}$, which degenerates to identity matrix $\mathbf{I}_i^k$. Similarly, the calculation of node information score in Equation (\ref{eq:rank}) can also be simplified as below:
\begin{equation}
	\mathbf{p} = \gamma(\mathcal{G}_i) = \Vert(\mathbf{I}_i^k - \mathbf{S}_i^k)\mathbf{H}_i^k\Vert_1,
\end{equation}
which makes our model very easy to implement.

\subsection{The Readout Function and Output Layer}
As we have demonstrated in Figure \ref{fig:model}, the neural network architecture repeats the graph convolution and pooling operations for several times, thus we would observe multiple subgraphs with different size in each level: $\mathbf{H}_i^1,\mathbf{H}_i^2,\cdots,\mathbf{H}_i^K$. To generate a fixed size graph level representation, we devise a readout function that aggregates all the node representations in the subgraph. Here, we simply use the concatenation of mean-pooling and max-pooling in each subgraph as follows:
\begin{equation}
	\mathbf{r}_i^k = \mathcal{R}(\mathbf{H}_i^k) = \sigma(\frac{1}{n_i^k}\sum_{p=1}^{n_i^k}\mathbf{H}_i^k(p,:)||\max_{q=1}^{d}\mathbf{H}_i^k(:,q)),
\end{equation}
where $\sigma(\cdot)$ is a nonlinear activation function and $\mathbf{r}_i^k \in \mathbb{R}^{2d}$. We then add\footnote{In our experiment, we use fixed size node representation across all layers, i.e., $d_{k} = \cdots = d_1 = d = 128$.} the readout outputs of different levels to form our final graph level representation:
\begin{equation}
	\mathbf{z}_i = \mathbf{r}_i^1 + \mathbf{r}_i^2 + \cdots + \mathbf{r}_i^K,
\end{equation}
which summarizes different levels' graph representations. 

Finally, we feed the graph level representation into MLP layer with softmax classifier, and the loss function is defined as the cross-entropy of predictions over the labels:
\begin{eqnarray}
	&\hat{\mathbf{Y}} = {\rm softmax}({\rm MLP}(\mathbf{Z})) \nonumber \\ 
	&\mathcal{L} = -\sum_{i \in L}\sum_{j=1}^c\mathbf{Y}_{ij}{\rm log}\hat{\mathbf{Y}}_{ij},
\end{eqnarray}
where $\hat{\mathbf{Y}}_{ij}$ represents the predicted probability that graph $\mathcal{G}_i$ belongs to class $j$, and $\mathbf{Y}_{ij}$ is the ground truth. $L$ denotes the training set of graphs that have labels.

\section{Experiments and Analysis}
\begin{table}[t]
    \centering
    \small
    \resizebox{.95\columnwidth}{!}
    {
    	\begin{tabular}{ c|c|c|c|c|c }
    	\hline
    	Datasets  	& $\#|\mathcal{G}|$ & $\#|\mathcal{V}|$	& Avg.$|\mathcal{V}|$ 	& Avg.$|\mathcal{E}|$ 	& $\#|c|$	\\
    	\hline	
    	ENZYMES     &   600   	  		& 19,580			&  32.63				& 62.14    				& 6 		\\
    	PROTEINS    &   1,113     		& 43,471			&  39.06				& 72.82   				& 2 		\\
    	D\&D		&	1,178	 		& 334,925			&  284.32				& 715.66				& 2 		\\
    	NCI1     	&   4,110    		& 122,747			&  29.87				& 32.30   				& 2 		\\
    	NCI109		&	4,127			& 122,494			&  29.68				& 32.13					& 2 		\\
   		Mutagenicity&	4,337			& 131,488			&  30.32  			 	& 30.77					& 2			\\
    	\hline
    	\end{tabular}
    }
    \caption{Statistics of the datasets.}
    \label{tab:datasets}
\end{table}

\begin{table*}
    \centering
    \small
    \begin{tabular}{ llcccccc }
    \toprule[0.8pt]
    Categories 				& Baselines & ENZYMES 	& PROTEINS	& D\&D 		& NCI1 		& NCI109 	& Mutagenicity		\\
    \cmidrule[0.5pt]{1-8}
    \multirow{3}{*}{Kernels}& GRAPHLET  & $29.16\pm5.63$ & $72.23\pm4.49$ & $72.54\pm3.83$ & $62.48\pm2.11$ & $60.96\pm2.37$ 						& $56.65\pm1.74$\\
    						& SP		& $42.66\pm5.38$ & $75.71\pm2.73$ & $78.72\pm3.89$ & $67.44\pm2.76$ & $67.72\pm2.28$ & $71.63\pm2.19$\\
    						& WL		& $51.16\pm6.19$ & $76.16\pm3.99$ & $76.44\pm2.35$ & $76.65\pm1.99$ & $76.19\pm2.45$ & $80.32\pm1.71$\\
    \cmidrule[0.5pt]{1-8}
    \multirow{3}{*}{GNNs}	& GCN     	& $43.66\pm3.39$ & $75.17\pm3.63$ & $73.26\pm4.46$ & $76.29\pm1.79$ & $75.91\pm1.84$ 						& $79.81\pm1.58$\\
    						& GraphSAGE	& $37.99\pm3.71$ & $74.01\pm4.27$ & $75.78\pm3.91$ & $74.73\pm1.34$ & $74.17\pm2.89$ & $78.75\pm1.18$\\
   							& GAT		& $39.83\pm3.68$ & $74.72\pm4.01$ & $77.30\pm3.68$ & $74.90\pm1.72$ & $75.81\pm2.68$ & $78.89\pm2.05$\\
   	\cmidrule[0.5pt]{1-8}
   	\multirow{7}{*}{Pooling}& Set2Set	& $33.16\pm3.21$ & $79.33\pm0.84$ & $70.83\pm0.84$ & $69.62\pm1.32$ & $73.66\pm1.69$ 						& $80.84\pm0.67$\\
   							& DGCNN		& $32.16\pm3.87$ & $79.99\pm0.44$ & $70.06\pm1.21$ & $74.08\pm2.19$	& $78.23\pm1.31$ & $80.41\pm1.02$ \\
   							& DiffPool	& $60.61\pm3.94$ & $79.90\pm2.95$ & $78.61\pm1.32$ & $77.73\pm0.83$ & $77.13\pm1.49$ & $80.78\pm1.12$ \\
   							& EigenPool & $63.97\pm2.51$ & $78.84\pm1.06$ & $78.63\pm1.36$ & $77.24\pm0.96$ & $75.99\pm1.42$ & $80.11\pm0.73$ 	\\
   							& gPool		& $43.33\pm2.88$ & $80.71\pm1.75$ & $77.02\pm1.32$ & $76.25\pm1.39$ & $76.61\pm1.39$ & $80.30\pm1.54$	\\
   							& SAGPool	& $43.99\pm4.23$ & $81.72\pm2.19$ & $78.70\pm2.29$ & $77.88\pm1.59$ & $75.74\pm1.47$ & $79.72\pm0.79$	\\
    						& EdgePool	& $65.33\pm4.36$ & $82.38\pm0.82$ & $79.20\pm2.61$ & $76.56\pm1.01$ & $79.02\pm1.89$ & $81.41\pm0.88$	\\
    \cmidrule[0.5pt]{1-8}
    \multirow{4}{*}{Proposed} 	& $\text{HGP-SL}_{{\rm NSL}}$	& $60.18\pm2.43$	& $81.51\pm1.69$	& $77.24\pm1.09$  	& $76.33\pm1.43$		& $76.32\pm1.22$								& $79.42\pm0.58$			\\
    							& $\text{HGP-SL}_{{\rm HOP}}$	& $62.16\pm2.11$	& $83.03\pm1.74$	& $78.42\pm1.37$  	& $77.72\pm1.54$			& $78.78\pm1.09$	& $79.88\pm1.09$			\\
    							& $\text{HGP-SL}_{{\rm DEN}}$	& $63.51\pm2.64$	& $83.12\pm0.84$	& $78.11\pm1.35$  	& $77.42\pm1.23$			& $78.76\pm0.61$	& $81.07\pm1.02$			\\
    							& HGP-SL	& \textbf{68.79 $\pm$ 2.11} & \textbf{84.91 $\pm$ 1.62} & \textbf{80.96 $\pm$ 1.26} & \textbf{78.45 $\pm$ 0.77}	& \textbf{80.67 $\pm$ 1.16} & \textbf{82.15 $\pm$ 0.58}	\\
    \bottomrule[0.8pt]
    \end{tabular}
    \caption{Graph classification in terms of accuracy with standard deviation (in percentage). We use \textbf{bold} to highlight wins.}
    \label{tab:classification_result}
\end{table*}

\subsection{Datasets}
We adopt six commonly used public benchmarks\footnote{Benchmarks are publicly available at https://ls11-www.cs.tu-dortmund.de/staff/morris/graphkerneldatasets} for empirical studies. Statistics of the six datasets are summarized in Table \ref{tab:datasets} with more descriptions as follows: \textbf{ENZYMES} \cite{borgwardt2005protein} is a dataset of protein tertiary structures, and each enzyme belongs to one of the 6 EC top-level classes. \textbf{PROTEINS} and \textbf{D\&D} \cite{dobson2003distinguishing} are two protein graph datasets, where nodes represent the amino acids and two nodes are connected by an edge if they are less than 6 Angstroms apart. The label indicates whether or not a protein is a non-enzyme. \textbf{NCI1} and \textbf{NCI109} \cite{shervashidze2011weisfeiler} are two biological datasets screened for activity against non-small cell lung cancer and ovarian cancer cell lines, where each graph is a chemical compound with nodes and edges representing atoms and chemical bonds, respectively. \textbf{Mutagenicity} \cite{kazius2005derivation} is a chemical compound dataset of drugs, which can be categorized into two classes: mutagen and non-mutagen.

\subsection{Baselines}
\subsubsection{Graph Kernel Methods.}
This group of methods perform graph classification by utilizing carefully designed kernels. We choose three classical algorithms: GRAPHLET \cite{shervashidze2009efficient}, Shortest-Path Kernel (SP) \cite{borgwardt2005shortest} and Weisfeiler-Lehman Kernel (WL) \cite{shervashidze2011weisfeiler} as baselines.
\subsubsection{Graph Neural Networks.}
Approaches in this group include representative graph neural networks: GCN \cite{kipf2016semi}, GraphSAGE \cite{hamilton2017inductive} and GAT \cite{velivckovic2017graph}, which are designed to learn meaningful node level representations. Therefore, we employ our proposed readout function to summarize the node representations for graph classification.
\subsubsection{Graph Pooling Models.}
In this group, we further consider numerous models that combine GNNs with pooling operator for graph level representation learning. Set2Set \cite{vinyals2015order} and DGCNN \cite{zhang2018end} are two novel global graph pooling algorithms. Another five hierarchical graph pooling models including DiffPool \cite{ying2018hierarchical}, gPool \cite{gao2019graph}, SAGPool \cite{lee2019self}, EdgePool \cite{diehl2019edge} and EigenPool \cite{ma2019graph} are also compared as baselines. 
\subsubsection{HGP-SL Variants.}
To further analyze the effectiveness of our proposed HGP-SL operator, we consider four variants here: $\text{HGP-SL}_{{\rm NSL}}$ (\textbf{N}o \textbf{S}tructure \textbf{L}earning) which discards the structure learning layer to verify the effectiveness of our proposed structure learning module, $\text{HGP-SL}_{{\rm HOP}}$ which removes the structure learning layer and connects the nodes within its $h$-hops, $\text{HGP-SL}_{{\rm DEN}}$ (\textbf{DEN}se) which employs the structure learning layer to learn a dense graph structure with softmax function defined in Equation (\ref{eq:softmax}) and HGP-SL which utilizes sparsemax function define in Equation (\ref{eq:sparse_max}) to learn a sparse graph structure. Both $\text{HGP-SL}_{{\rm DEN}}$ and HGP-SL use efficiency improved structure learning strategy.
\subsubsection{Experiment and Parameter Settings.}
Following many previous work \cite{ying2018hierarchical,ma2019graph}, we randomly split each dataset into three parts: $80\%$ as training set, $10\%$ as validation set and the remaining $10\%$ as test set. We repeat this randomly splitting process 10 times, and the average performance with standard derivation is reported. For baseline algorithms, we use the source code released by the authors, and their hyper-parameters are tuned to be optimal based on the validation set. In order to ensure a fair comparison, the same neural network architectures are used for the existing pooling baselines and our proposed model. The dimension of node representations is set as 128 for all methods and datasets. We implement our proposed HGP-SL with PyTorch, and the Adam optimizer is utilized to optimize the model. The learning rate and weight decays are searched in $\{0.1,0.01,0.001,1e^{-4},1e^{-5}\}$, pooling ratio $r \in [0.1,0.9]$ and layers $K \in [1,5]$. The MLP consists of three fully connected layers with number of neurons in each layer setting as 256, 128, 64, followed by a softmax classifier. Early stopping criterion is employed in the training process, i.e., we stop training if the validation loss dose not decrease for 100 consecutive epochs. The source code is publicly available\footnote{Code is available at https://github.com/cszhangzhen/HGP-SL}.

\subsection{Performance on Graph Classification}
The classification performance is reported in Table \ref{tab:classification_result}. To summarize, we have the following observations:
\begin{itemize}
	\item First of all, a general observation we can draw from the results is that our proposed HGP-SL consistently outperforms other state-of-the-art baselines among all datasets. For instance, our method achieves about 3.08\% improvement over the best baseline in PROTEINS dataset, which is 12.97\% improvement over GCN with no hierarchical pooling mechanism. This verifies the necessity of adding graph pooling module.
	\item It is worth noting that the traditional graph kernel based methods demonstrate competitive performance. However, the carefully designed graph kernels typically involve massive human domain knowledge, which has difficulty in generalizing to graphs with arbitrary structures. Furthermore, the two-stage procedure of extracting graph features and performing graph classification might result in sub-optimal performance.
	\item Being consistent with previous work's findings \cite{ma2019graph}, we also observe that the GNNs group can not achieve satisfied results. We argue that the major reason is because they ignore the graph structure information when globally summarizing the node representations, which further verifies the necessity of adding graph pooling module.
	\item In particular, the global pooling approaches Set2Set and DGCNN are surpassed by most of the hierarchical pooling methods with a few exceptions. This is because their learned graph representations are still ``flat", and the hierarchical structure information or functional units in the graph are ignored, which play an important role in predicting the entire graph labels.
	\item We note that the hierarchical pooling models can achieve relative better performance among most baselines, which further shows the effectiveness of the hierarchical pooling mechanism. Among them, gPool and SAGPool perform poorly in ENZYMES dataset. This may be due to the limited training samples per class resulting in the neural network overfitting. EdgePool gains superior performance in this group of competitors, which scales down the size of graphs by contracting each pair of nodes in the graph. Obviously, our proposed HGP-SL outperforms EdgePool with different gains for all settings.
	\item Finally, HGP-SL and $\text{HGP-SL}_{{\rm DEN}}$ obtain better performance than $\text{HGP-SL}_{{\rm NSL}}$ and $\text{HGP-SL}_{{\rm HOP}}$, which justifies the effectiveness of our proposed structure learning layer. Moreover, $\text{HGP-SL}_{{\rm HOP}}$ performs worse than HGP-SL. This is because the disconnected nodes are still unreachable in its $h$-hops. HGP-SL further outperforms $\text{HGP-SL}_{{\rm DEN}}$, which indicates the learned dense graph structure might introduce additional noisy information and degenerate the performance. Furthermore, in the real-world scenario, graphs usually have sparse topologies, thus our proposed HGP-SL could learn more reasonable graph structures compared with $\text{HGP-SL}_{{\rm DEN}}$.
\end{itemize}

\subsection{Ablation Study and Visualization}
\subsubsection{HGP-SL Convolutional Neural Network Architectures.}
As mentioned in previous sections, our proposed HGP-SL can be integrated into various graph neural network architectures. We consider three most widely used graph convolutional architectures as our model's building block to investigate the affect of different convolution operations: GCN \cite{kipf2016semi}, GraphSAGE \cite{hamilton2017inductive} and GAT \cite{velivckovic2017graph}. We evaluate them on three datasets, which cover both small and large datasets. Their results are shown in Table \ref{tab:result_architecture}. Similar results can also be found in the remaining datasets, thus we omit them due to the limited space. As demonstrated in Table \ref{tab:result_architecture}, the performance on graph classification varies depending on which dataset and the type of GNN in HGP-SL are chosen. In addition, we also combine the top-K selection procedure proposed in gPool and SAGPool with our proposed structure learning. We name them as gPool-SL and SAGPool-SL for short. From the results, we observe that gPool-SL and SAGPool-SL outperform gPool and SAGPool by incorporating the structure learning mechanism, which verifies the effectiveness of our proposed structure learning.
\subsubsection{Hyper-parameter Analysis.}
\begin{figure}
  \centering
  \includegraphics[width=\columnwidth]{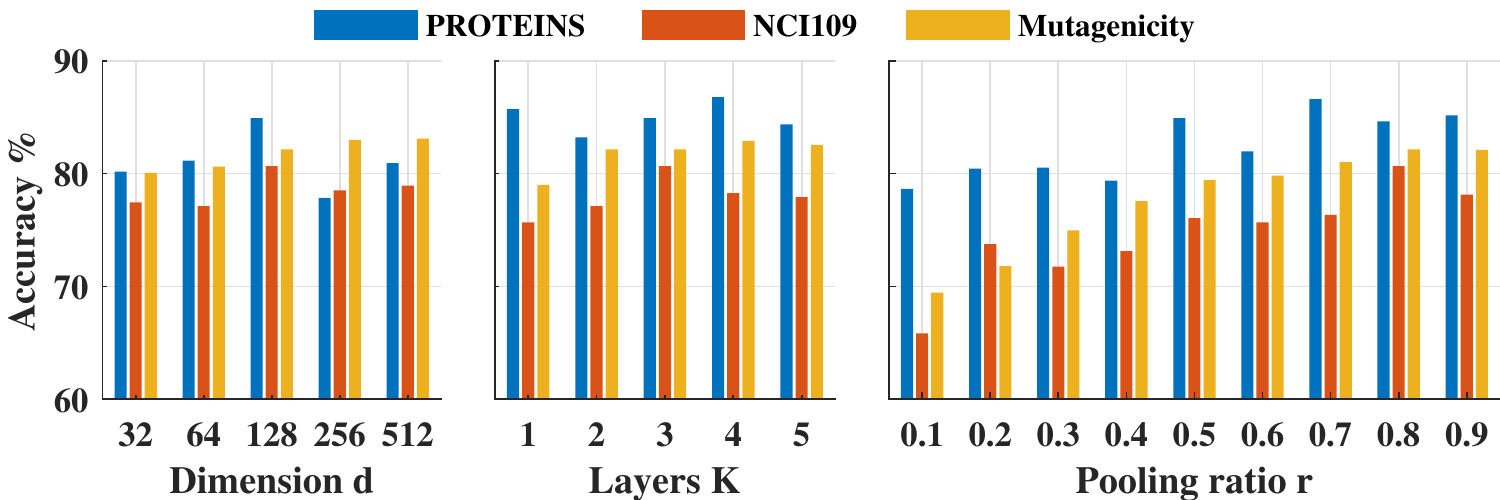}
  \caption{Hyper-parameter sensitivity analysis.}
  \label{fig:hyper_parameter}
\end{figure}

We further study the sensitivities of several key hyper-parameters by varying them in different scales. Specifically, we investigate how the number of neural network layers $K$, graph representation dimension $d$ and pooling ratio $r$ will affect the graph classification performance. As we can see in Figure \ref{fig:hyper_parameter}, HGP-SL almost achieves the best performance across different datasets when setting $K = 3$, $d = 128$ and $r=0.8$, respectively. The pooling ratio $r$ cannot be too small, otherwise most of the graph structure information will be lost during the pooling process. 
\subsubsection{Visualization.}
\begin{figure}[!htb]\small
\centering
\subfigure[\scriptsize{The Original Graph}]{
\label{fig:example1}
\centering
\includegraphics[width=0.13\textwidth]{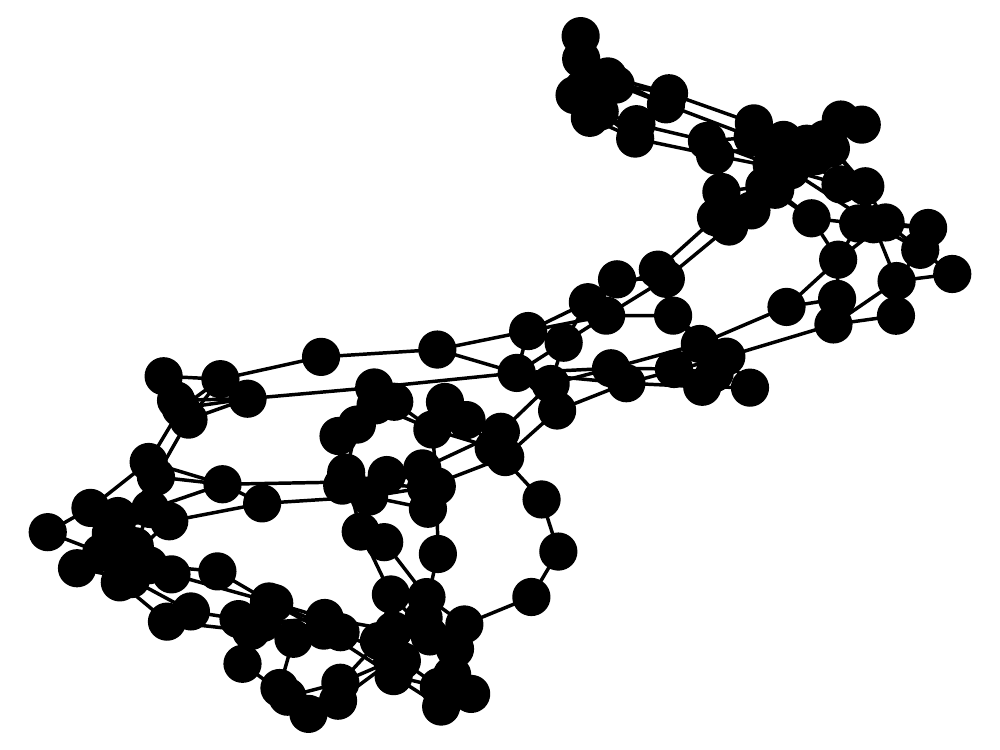}
}
\subfigure[\scriptsize{gPool Pool3}]{
\label{fig:example2}
\centering
\includegraphics[width=0.13\textwidth]{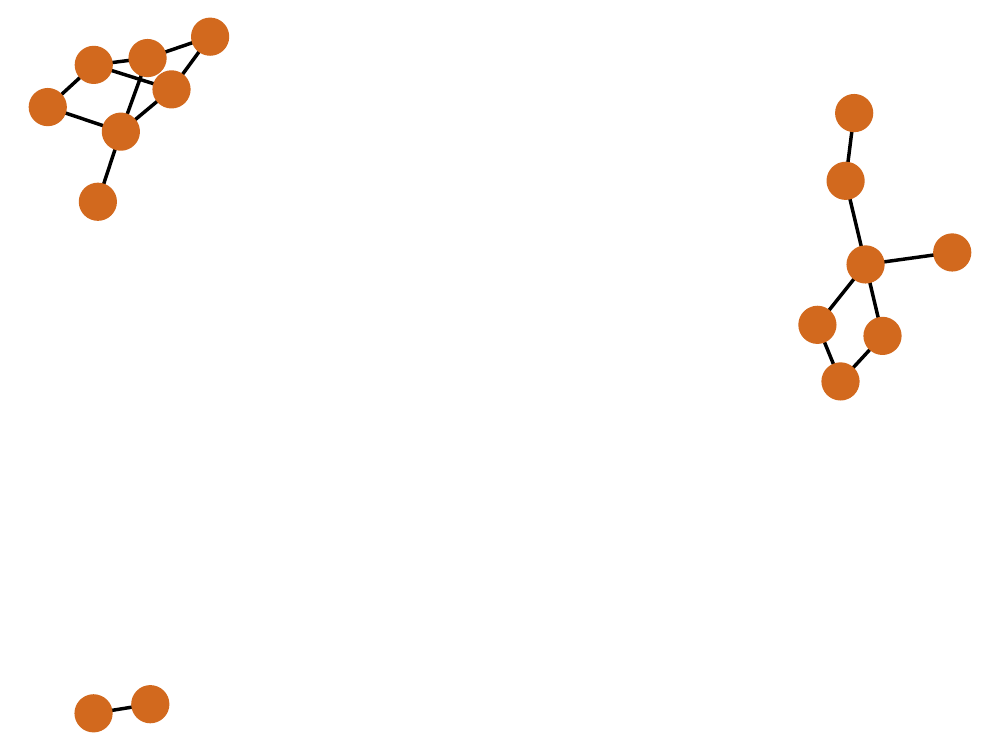}
}
\subfigure[\scriptsize{SAGPool Pool3}]{
\label{fig:example3}
\centering
\includegraphics[width=0.13\textwidth]{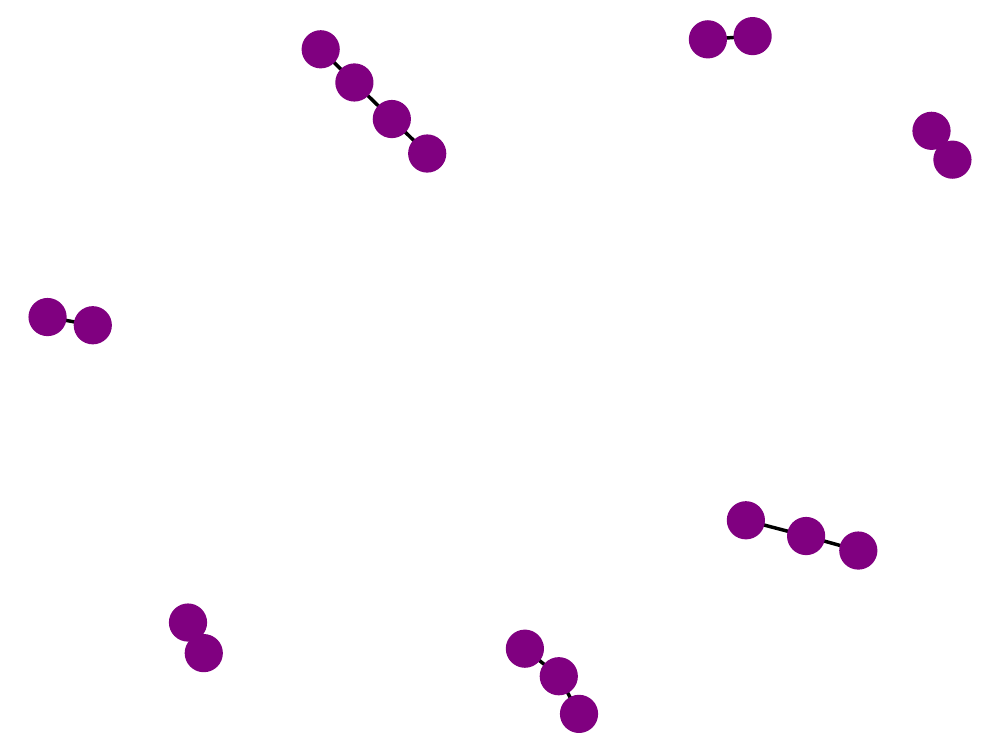}
}
\subfigure[\scriptsize{$\text{HGP-SL}_{{\rm NSL}}$ Pool3}]{
\label{fig:example4}
\centering
\includegraphics[width=0.13\textwidth]{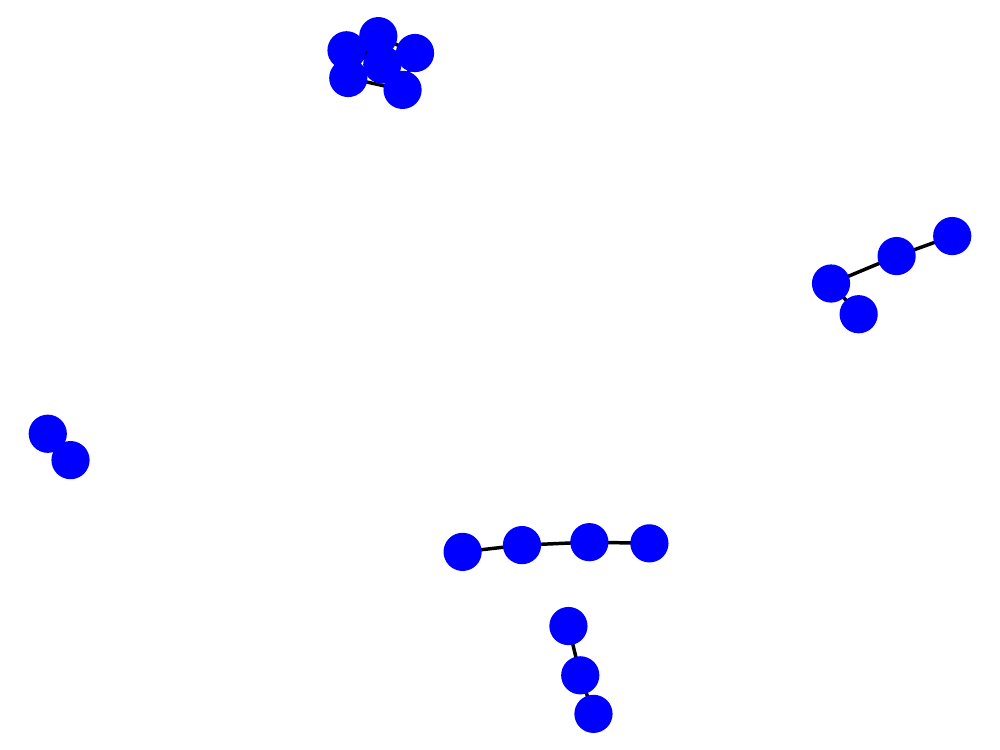}
}
\subfigure[\scriptsize{$\text{HGP-SL}_{{\rm DEN}}$ Pool3}]{
\label{fig:example5}
\centering
\includegraphics[width=0.13\textwidth]{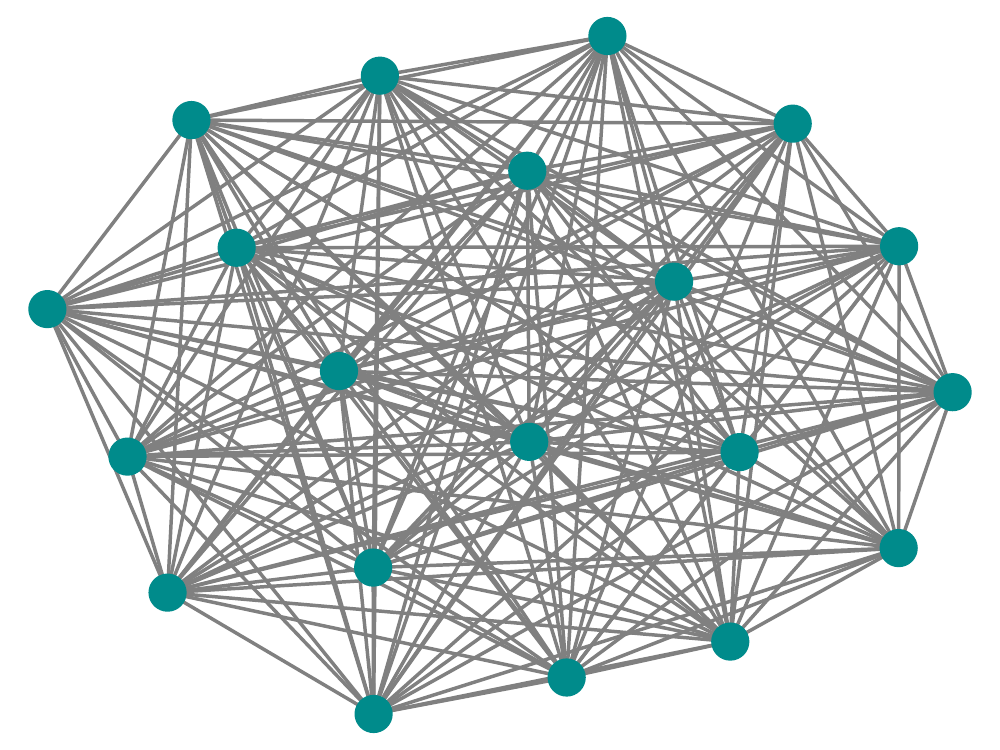}
}
\subfigure[\scriptsize{HGP-SL Pool3}]{
\label{fig:example6}
\centering
\includegraphics[width=0.13\textwidth]{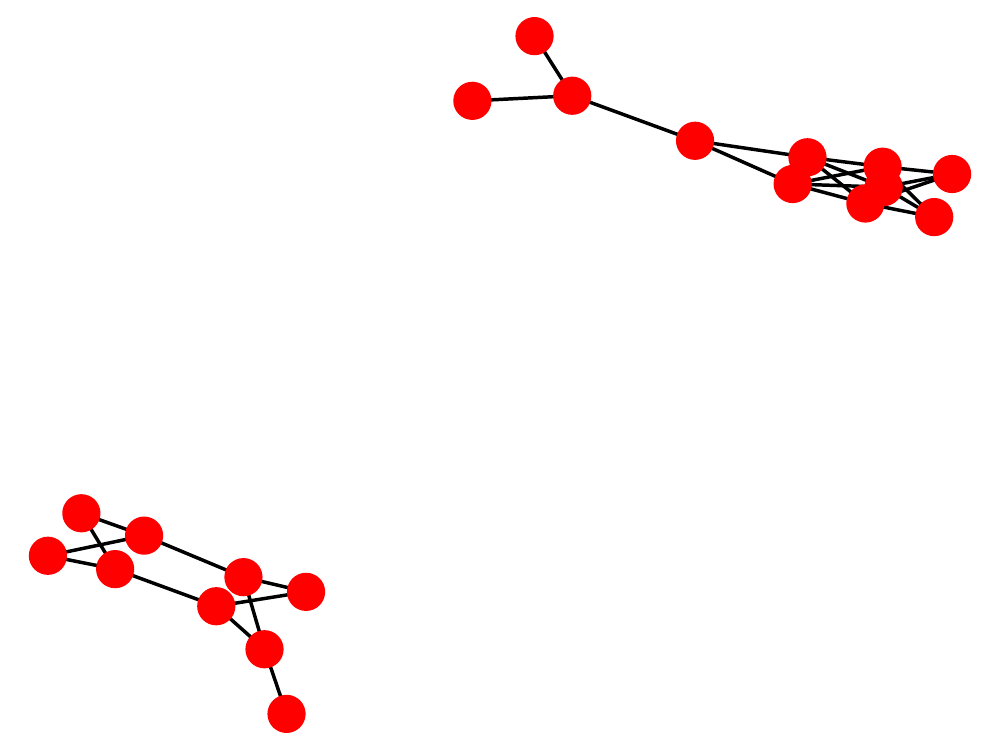}
}
\caption{Visualization of different pooling methods.}
\label{fig:visualization}
\end{figure}

We utilize networkx\footnote{https://networkx.github.io/} to visualize the pooling results of HGP-SL and its variants. In detail, we randomly sample a graph from PROTEINS dataset, which contains 154 nodes. We build a three layer graph neural network with pooling ratio setting as 0.5, which then generates three pooled graphs with nodes as 77, 39 and 20 respectively. We plot the 3rd pooled graph in Figure \ref{fig:visualization}. It shows $\text{HGP-SL}_{{\rm NSL}}$ and $\text{HGP-SL}_{{\rm DEN}}$ fail to preserve meaningful graph topologies, while HGP-SL is able to preserve relatively reasonable topology of the original protein graph after pooling.

\begin{table}[t]
    \centering
    \small
    \begin{tabular}{lccc}
    \hline
    Architectures  				  	& PROTEINS			& NCI109			& Mutagenicity		\\
    \hline	
    $\text{HGP-SL}_{{\rm GCN}}$  	& 84.91$\pm$1.62 	& 80.67$\pm$1.16	& 82.15$\pm$0.58 	\\
    $\text{HGP-SL}_{{\rm GAT}}$  	& 85.04$\pm$1.01	& 79.82$\pm$1.06	& 82.02$\pm$0.81	\\
    $\text{HGP-SL}_{{\rm SAGE}}$ 	& 84.99$\pm$0.82    & 80.11$\pm$0.96 	& 81.96$\pm$0.97	\\
    \hline
    $\text{gPool-SL}$ 				& 81.25$\pm$1.27    & 77.71$\pm$1.22 	& 80.42$\pm$1.08	\\
    $\text{SAGPool-SL}$ 			& 82.67$\pm$1.42 	& 78.01$\pm$1.50	& 80.00$\pm$1.22	\\
    \hline
    \end{tabular}
    \caption{HGP-SL performance with different architectures.}
    \label{tab:result_architecture}
\end{table}

\section{Conclusion}
In this paper, we investigate graph level representation learning for the task of graph classification. We propose a novel graph pooling operator HGP-SL, which empowers GNNs to learn hierarchical graph representations. It can also be conveniently integrated into various GNN architectures. Specifically, the graph pooling operation is a non-parametric step, which utilizes node features and graph structure information to perform down-sampling on graphs. Then, a structure learning layer is stacked on the pooling operation, which aims to learn a refined graph structure that can best preserve the essential topological information. We combine the proposed HGP-SL operator with graph convolutional neural networks to conduct graph classification task. Comprehensive experiments on six widely used benchmarks demonstrate its superiority to a range of state-of-the-art methods.

\bibliographystyle{aaai}
\bibliography{reference}

\newpage
\section{Appendix}

\subsection{Proof for Algorithm 1}
To summarize, ${\rm sparsemax}(\cdot)$ considers the Euclidean projection of the input vector $\mathbf{z}$ onto the probability simplex, which can be defined as the following optimization problem:
\begin{eqnarray}
	&\min_{\mathbf{p} \in \mathbb{R}^n} & \frac{1}{2} \Vert \mathbf{p} - \mathbf{z} \Vert^2 \nonumber \\
	&{\rm s.t.}& \mathbf{p}^{\top}\mathbf{1} = 1, \quad \mathbf{p} \geq 0.
	\label{eq:projection}
\end{eqnarray}
Then, the Lagrangian of the optimization problem in Equation (\ref{eq:projection}) is:
\begin{equation}
	\mathcal{L}(\mathbf{p},\bm{\alpha},\beta) = \frac{1}{2}\Vert \mathbf{p} - \mathbf{z} \Vert^2 - \bm{\alpha}^{\top}\mathbf{p} + \beta(\mathbf{1}^{\top}\mathbf{p} - 1).
	\label{eq:lagrangian}
\end{equation}
The optimal $(\mathbf{p}^*, \bm{\alpha}^*, {\beta}^*)$ must satisfy the following Karush-Kuhn-Tucker conditions:
\begin{equation}
	\mathbf{p}^* - \mathbf{z} - \bm{\alpha}^* + {\beta}^*\mathbf{1} = \mathbf{0},
	\label{eq:condition1}
\end{equation}
\begin{equation}
	\mathbf{1}^{\top}\mathbf{p}^* = 1, \quad \mathbf{p}^* \geq \mathbf{0}, \quad \bm{\alpha}^* \geq \mathbf{0},
	\label{eq:condition2}
\end{equation}
\begin{equation}
	\alpha_i^*p_i^* = 0, \quad \forall i \in \{1, \cdots, n\}.
	\label{eq:condition3}
\end{equation}
If for $\forall i \in \{1, \cdots, n\}$ we have $p_i^* > 0$, then from Equation (\ref{eq:condition3}) we must satisfy $\alpha_i^* = 0$. Thus, from Equation (\ref{eq:condition1}) we can get $p_i^* = z_i - \beta^*$. Let $S(\mathbf{z}) = \{j \in \{1,\cdots,n\}|p_j^*>0\}$. From Equation (\ref{eq:condition2}) we obtain $\sum_{j\in S(\mathbf{z})}(z_j-\beta^*) = 1$, which yields the Line 3 in Algorithm 1, i.e., $\beta^*=\tau(\mathbf{z})$. Again from Equation (\ref{eq:condition3}), we have that $\alpha_i^* >0$ implies $p_i^*=0$, which from Equation (\ref{eq:condition1}) implies $\alpha_i^* = \beta^* - z_i \geq 0$, i.e., $z_i \leq \beta^*$ for $i \notin S(\mathbf{z})$. Thus, we have the procedure in Algorithm 1.
\end{document}